%% file: acl2020.tex
\documentclass[11pt,a4paper]{article}
\usepackage[hyperref]{acl2020}
\usepackage{times}
\usepackage{latexsym}

\usepackage{microtype}

\usepackage{graphicx}
\usepackage{amsmath}
\usepackage{makecell}
\usepackage{booktabs}
\usepackage{multirow}
\usepackage[normalem]{ulem}
\usepackage{cancel}
\usepackage{cleveref}
\usepackage{enumitem}
\usepackage{tikz}
\usetikzlibrary{tikzmark}
\usepackage{hyperref}
\usepackage[none]{hyphenat}
\usepackage{fancyvrb}
\crefname{section}{§}{§§}
\Crefname{section}{§}{§§}
\crefformat{footnote}{#2\footnotemark[#1]#3} 

\usepackage{url}

\aclfinalcopy 

\title{Evaluating Pretrained Transformer Models for Entity Linking in Task-Oriented Dialog}

\author{
    Sai Muralidhar Jayanthi,
    Varsha Embar,
    Karthik Raghunathan \\
    MindMeld, Cisco Systems \\
    \texttt{\{saijayan, vembar, ktick\}@cisco.com}
}

\date{}

\begin{document}
\maketitle

\input{sections/abstract}

\input{sections/introduction}
\input{sections/models}
\input{sections/datasets}
\input{sections/analysis}

\input{sections/conclusion}

\bibliography{anthology,custom,custom2}
\bibliographystyle{acl_natbib}

\appendix

\newpage

\section{Appendices}
\label{sec:appendix}

\input{sections/appendix-datasets}
\input{sections/appendix-experiments}

\end{document}

%% file: sections/abstract.tex
\begin{abstract}
The wide applicability of pretrained transformer models (PTMs) for natural language tasks is well demonstrated, but their ability to comprehend short phrases of text is less explored. To this end, we evaluate different PTMs from the lens of unsupervised Entity Linking in task-oriented dialog across 5 characteristics-- \textit{syntactic}, \textit{semantic}, \textit{short-forms}, \textit{numeric} and \textit{phonetic}. Our results demonstrate that several of the PTMs produce sub-par results when compared to traditional techniques, albeit competitive to other neural baselines. We find that some of their shortcomings can be addressed by using PTMs fine-tuned for text-similarity tasks, which illustrate an improved ability in comprehending semantic and syntactic correspondences, as well as some improvements for short-forms, numeric and phonetic variations in entity mentions. We perform qualitative analysis to understand nuances in their predictions and discuss scope for further improvements.\footnote{Code and re-purposed datasets can be found at \href{https://github.com/murali1996/el\_tod}{https://github.com/murali1996/el\_tod}}
\end{abstract}

%% file: sections/introduction.tex
\section{Introduction}
\label{sec:introduction}

In task-oriented dialog systems, Entity Linking (EL) is the process of
disambiguating a detected entity mention (\textit{aka.} slot)
in a user utterance to a canonical
entry in a Knowledge Base (KB).
EL is a crucial step in building 
robust dialog systems, especially 
when dealing with domain-specific 
entities, \textit{e.g.}, a chatbot for food 
ordering or a voice assistant for 
medical assistance.

Popular open-source conversational AI platforms such as DeepPavlov \cite{burtsev-etal-2018-deeppavlov}, MindMeld \cite{raghuvanshi-etal-2018-developing} and Rasa \cite{Bocklisch2017RasaOS} maintain a KB of canonical entries, each consisting of a title, optionally with aliases  (\textit{i.e.}, alternate usages)
for the task of entity linking.
Detected entities from user utterances,
often with spelling and automatic speech recognition (ASR) errors, are then
mapped to those canonical 
entries through text classification
or similarity matching techniques.\footnote{
Entity Linking may be clubbed with 
Entity Recognition
or is a standalone component of the NLP pipeline, 
the latter is used in this work for better 
interpretability.
}
\input{images/extended_abstract_examples}

Previous works 
\cite{chen-etal-2020-contextualized, 
decao2020autoregressive, 
broscheit-2019-investigating}
have proposed
\textit{context-aware} classification techniques for EL,
wherein the context 
surrounding the slots is leveraged to
ascertain canonical names.
However, 
such approaches 
fall short
due to 
\textbf{(i)} 
their reliance on 
large training/fine-tuning sets 
and associated annotation costs
\textbf{(ii)} 
requirement to re-train the classifiers
with every change in KB entries.
Alternatively, a more popular paradigm 
is to model EL as a matching problem 
by transforming entities into \textit{vectors},
and using a similarity function such as cosine distance 
to find the closest canonical entry.

EL systems
typically rely on 
textual n-gram features 
modeled by
ranking algorithms 
such as
BM25~\cite{10.5555/188490.188561} 
implemented as part
of search engines such as Elasticsearch.\footnote{
\href{https://www.elastic.co/blog/practical-bm25-part-2-the-bm25-algorithm-and-its-variables}{https://www.elastic.co/blog/practical-bm25}}
To capture semantic similarity 
within such systems,
one needs to 
tediously engineer feature 
sets and collect synonyms or 
aliases for each KB entry, leading
to a lot of manual effort and 
development cost.

Recently, pretrained word embeddings
have had much success in capturing entity
correspondences 
\cite{francis-landau-etal-2016-capturing, 
10.5555/2832415.2832435}
by addressing 
aforementioned
shortcomings-- 
off-the-shelf usage without reliance on training data
and flexibility to expand KBs without 
retraining.
\citet{Mudgal2018DeepLF} 
presents a detailed account 
of different deep learning based representations and modeling
choices for the EL task, showing the advantages of using them 
over traditional systems.
 
More recently, transformer-based 
PTMs like
BERT \cite{devlin-etal-2019-bert} 
have excelled 
for Entity Linking
when 
entities are in the 
form of tabular data
without much additional \textit{context} \cite{Tracz2020BERTbasedSL, 10.1145/3340531.3412131, 
10.14778/3421424.3421431, Mudgal2018DeepLF}.
However, 
their ability
to understand nuances 
in linking 
short spans of free-form text 
is not 
thoroughly tested,
especially for domain-specific 
entities with minimal context.

In this work,
we investigate and 
analyze how 
different PTMs
behave in
such
settings,
when compared to 
widely adopted 
neural and non-neural
models  
\textbf{(\cref{sec:models})}.
To probe model behaviours on examples 
with different characteristics, 
we curate and benchmark evaluation datasets 
of various sizes that each contain a
subset of those characteristics \textbf{(\cref{sec:datasets})}. 
Lastly, we present
qualitative 
as well as 
quantitative analysis
of the predictions 
of various 
models, which shows that 
while 
pretrained models 
fine-tuned for
text-similarity tasks 
perform the best overall,
there is room for improvement
\textbf{(\cref{sec:analysis})}.

%% file: images/extended_abstract_examples.tex
\begin{figure}
\centering
\includegraphics[width=0.50\textwidth]{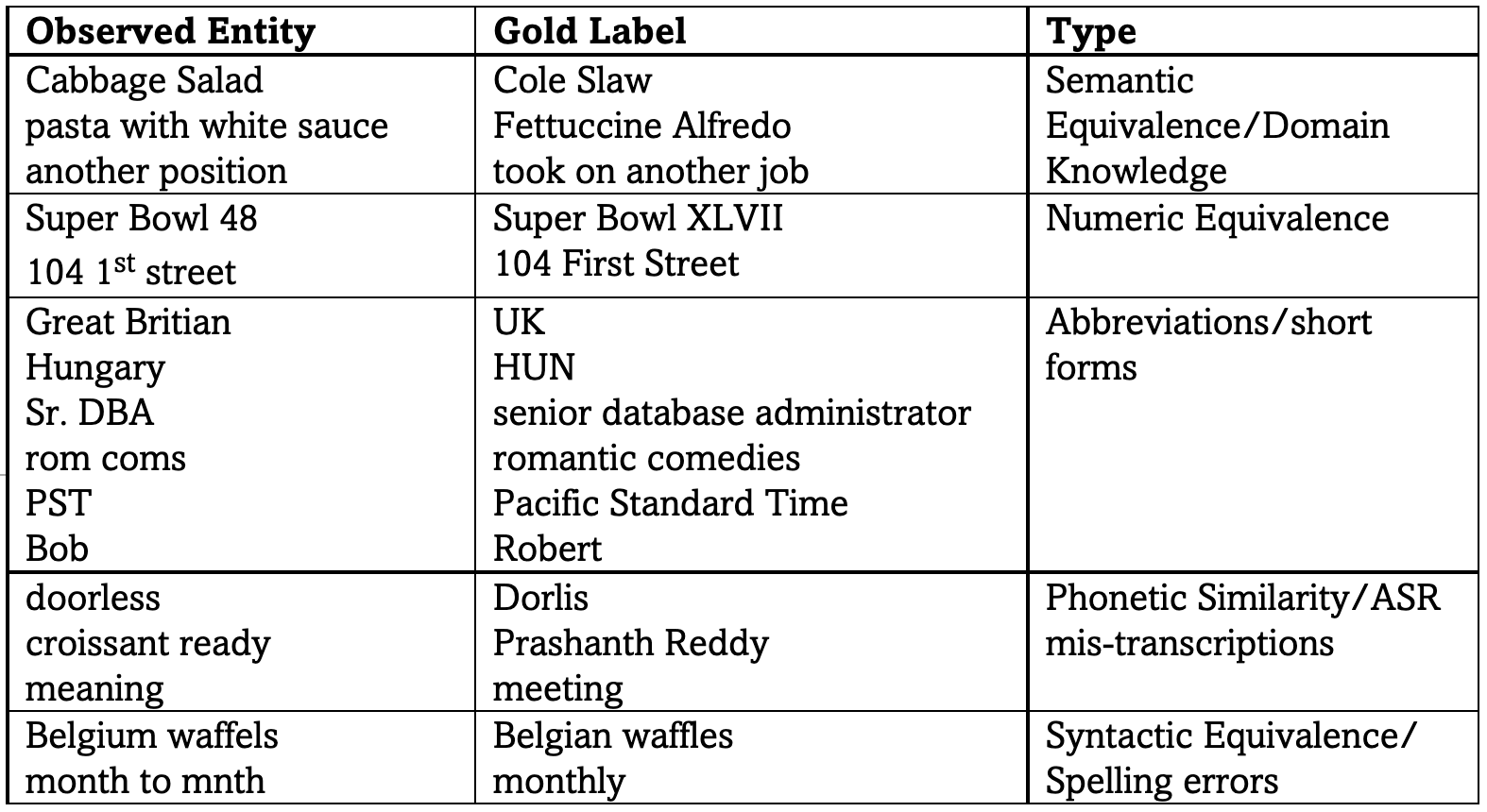}
\caption{
    \small Different types of matching scenarios observed in 
    Entity Linking task for short spoken/written language texts.
}
\label{fig:list_of_examples}
\end{figure}

%% file: sections/models.tex
\section{Models}
\label{sec:models}

In this section, we provide a brief detail of the different pretrained transformer models (PTMs) as well as the 5 baseline models (3 neural and 2 non-neural) used in our benchmarking process. We categorize PTMs under consideration into 4 different types to understand the usefulness of different pre-training strategies, number of parameters and inference times. We adopt the model nomenclature from Huggingface\footnote{\href{https://huggingface.co/models}{https://huggingface.co/models}} ~\cite{wolf-etal-2020-transformers} and refer the reader to \citet{Rogers_2020} and \citet{Qiu2020PretrainedMF} for more comprehensive account on these different types of PTMs and their utility.

We categorize the PTMs as follows:

\paragraph{Type-I} 
Pretrained general-purpose transformer language models which are \textit{base}-sized. These include \textit{bert-base-cased}~\cite{devlin-etal-2019-bert}, \textit{roberta-base}~\cite{Liu2019RoBERTaAR} and \textit{mpnet-base}~\cite{Song2020MPNetMA}.

\paragraph{Type-II} 
Parameters-reduced models which are also trained for language modeling tasks through different parameter reduction techniques. These include \textit{albert-base-v2}~\cite{Lan2020ALBERT}, \textit{distilbert-base-cased}~\cite{sanh2019distilbert}, \textit{distilroberta-base}~\cite{sanh2019distilbert}, and \textit{MiniLM-L6-uncased}~\cite{NEURIPS2020_3f5ee243}.

\paragraph{Type-III}
\citet{Reimers_2019} fine-tuned some of the Type-I and Type-II models on a variety of datasets annotated for textual similarity tasks\footnote{\href{https://www.sbert.net/docs/pretrained\_models.html}{https://www.sbert.net/docs/pretrained\_models.html}}. We select their \textit{all-*} models which were fine-tuned with more than 1 billion textual pairs and were designed as general purpose textual similarity models. These include \textit{all-distilroberta-v1}, \textit{all-mpnet-base-v2} and \textit{all-MiniLM-L6-v2}.

\paragraph{Type-IV}
Dynamic quantization can reduce the size of the model while only having a limited implication on accuracy. We use Pytorch's \cite{NEURIPS2019_9015} dynamic quantization functionality\footnote{\href{https://pytorch.org/tutorials/intermediate/dynamic\_quantization\_bert\_tutorial.html}{https://pytorch.org/dynamic\_quantization\_bert\_tutorial.html}} to obtain the quantized versions of the following models: \textit{all-mpnet-base-v2} and \textit{all-MiniLM-L6-v2}.

\paragraph{} In addition to the pretrained language models based on transformer architecture, we also benchmark PTMs based on other neural architectures. Specifically, we consider the following 3 neural  models as baselines--
(1) \textsc{Fasttext} \cite{Bojanowski2017EnrichingWV},
(2) \textsc{Flair} \cite{akbik2019flair},
and
(3) \textsc{Elmo} \cite{Peters2018DeepCW}.

\paragraph{\textsc{Fasttext}} consists of continuous distributed word representations trained on large unlabeled corpora for many natural language processing tasks. It represents each word as the sum of its  character n-grams. Compared to \textsc{Flair} and \textsc{Elmo}, this model has a shallower  network and is pretrained similar to \citet{Mikolov2013DistributedRO}'s skipgram model with negative sampling. In our benchmarking, we use the 300-dimension English model.\footnote{\href{https://github.com/facebookresearch/fastText/blob/master/docs/crawl-vectors.md}{https://github.com/facebookresearch/fastText/crawl-vectors.md}}

\paragraph{\textsc{Flair}} is a LSTM based pretrained character language model~\cite{10.1162/neco.1997.9.8.1735}, trained to produce a novel type of word embedding also known as \textit{contextual string embeddings}. It is trained without any explicit notion of words and hence can represent even out-of-vocabulary (OOV) words similar to \textsc{Fasttext}.
In our experiments, we use word representations concatenated from their \textit{news-forward} and \textit{news-backward} models leading to 4096-dimensional vectors.\footnote{\href{https://github.com/flairNLP/flair/blob/master/resources/docs/embeddings/FLAIR\_EMBEDDINGS.md}{https://github.com/flairNLP/flair/FLAIR\_EMBEDDINGS.md}}

\paragraph{\textsc{Elmo}} is a deep contextualized bidirectional word representation produced by pretrained LSTMs. In our experiments, we use the \textit{base} model and concatenate all three ELMo layers leading to 3072-dimensional vectors.\footnote{\href{https://github.com/allenai/bilm-tf}{https://github.com/allenai/bilm-tf}}

\paragraph{} We compare all the above neural models with two non-neural baselines which are popularly adopted for the task at hand-- (1) \textsc{tfidf} vectorizer\footnote{\href{https://tinyurl.com/sklearn-TFIDF}{https://scikit-learn.org/sklearn-TFIDF}} and 
(2) \textsc{bm25}, both using word 
\& character n-grams upto 5-gram.

For all models except \textsc{bm25},
we use cosine similarity as the
scoring function. 
For every pretrained model,
we use mean pooled representation
of all (sub-)words
in a given entity text 
as its final representation.\footnote{Different pretrained models have different tokenization strategies and we leave any analysis on the effect of tokenization to future work.}

%% file: sections/datasets.tex
\section{Datasets}
\label{sec:datasets}
We utilize both in-house and publicly available corpora to curate datasets in English for the Entity Linking task-- MindMeld Blueprints dataset\footnote{\href{https://github.com/CiscoDevNet/mindmeld-blueprints}{https://github.com/CiscoDevNet/mindmeld-blueprints}}(\textsc{mm\_bp}) along with word and character level misspelled versions of this data (\textsc{mm\_bp-word} and \textsc{mm\_bp-char}), re-purposed open-domain QA datasets like ComplexWebQuestions (\textsc{complwq}) \cite{talmor18compwebq} and MKQA (\textsc{mkqa}) \cite{mkqa}, acronym identification dataset (\textsc{acri})~\cite{veyseh-et-al-2020-what}, and an in-house dataset of ASR mis-transcriptions for person names (\textsc{asr-mis}).
More details on the dataset curation process is provided in Appendix A.

To probe model behaviours further, we manually annotate 
1.3K 
queries pooled from all of these datasets into our 5 predefined categories as follows (with their sample sizes) --
\textsc{semantic} (\#294), 
\textsc{syntactic} (\#408),
\textsc{short-forms} (\#310), 
\textsc{numerals} (\#125) 
and 
\textsc{phonetics} (\#200).
Examples from these sets are presented in Figure~\ref{fig:list_of_examples}.

We use Precision@1 (P@1) and Precision@5 (P@5) 
as our benchmarking metrics
and conduct all our experiments
using the publicly available MindMeld 
framework\footnote{\href{https://github.com/cisco/mindmeld}{https://github.com/cisco/mindmeld}}.
Unless otherwise stated,
we do not include any aliases alongside
canonical titles for matching KB entries 
and utilize all known 
aliases as our test queries.
We disregard any canonical descriptions
as they are not always available and procuring 
them may have significant annotation costs.

%% file: sections/analysis.tex
\section{Results \& Analysis}
\label{sec:analysis}

\input{tables/results_table}

Table~\ref{tab:results_table} 
present the results of
different models 
across our curated datasets.
We observe that on average,
Type-I 
\& 
Type-II models
perform poorly compared 
to the baselines
by atleast 30\% P@1. 
However, Type-III 
\& 
Type-IV 
models, 
 fine-tuned 
to find similar sentence pairs,
perform superior to our
baselines by 5-13\%,
showcasing the usefulness 
of such tuning strategies even to short texts.
We further observe that 
the parameter-reduced
models generally perform 
better than the \textit{base} 
models. Almost 
all 
PTMs 
perform poorly 
on abbreviations and also fail to beat the 
\textsc{bm25} 
baseline on the phonetic matching dataset.
While we believe that these two datasets are quite
challenging to the PTMs as their training processes
do not include any related objectives, the superior
performance of Type-III models compared to Type-I and Type-II
is quite encouraging.
On misspelled 
versions of the datasets, 
Type-III \& Type-IV models
still perform better than others. However, their precision falls short by at least 10\% absolute indicating scope for improvement.

\subsection{Qualitative analysis}
Figure \ref{fig:bar_plot} shows the performance of different models on the 5 different categories of data without and with 
aliases in the KB.
We perform a manual inspection of the results across the 5 categories with 3 different models: baseline BM25 model, Type-I \textit{bert-base-cased} (BERT) and Type-IV
\textit{all-mpnet-base-v2-quantized} (MPNET-Q).

\input{images/bar_plot}

\subsubsection{Syntactic Matches}
Syntactic matches refer to cases when the query and its matching canonical form have slight textual variations or spelling errors. The baseline TFIDF and BM25 models are well equipped to handle such differences and perform on-par and in some cases, better than the other models. Between the BERT and MPNET-Q models, the latter handles syntactic differences better than the former by favouring more word overlaps.
     
\begin{Verbatim}[fontsize=\small]
    Query: John Jr. 
    BM25: John F. Kennedy Jr. 
    BERT: Michael Joseph Jackson, Jr. 
    MPNET-Q: John Warner 
    
    Query: mammoth pizza 
    BM25: Wham, Bam, Thank You Mammoth 
    BERT: Pizzawich 
    MPNET-Q: Fresco Pizza
    
    Query: Hindi 
    BM25: Hindi Language 
    BERT: India 
    MPNET-Q: Hindi Language 
\end{Verbatim}

\subsubsection{Semantic Matches}
The baseline BM25 system relies heavily on aliases to handle queries that are semantically equivalent to one of the canonical names in the KB. In their absence, the model performs poorly in this category. In contrast, the transformer models are better suited to handle these queries. We notice 2 trends in the BERT and MPNET-Q models: \\
While BERT tends to predict related words, they are not always semantically equivalent. 
    
\begin{Verbatim}[fontsize=\small]    
    Query: Instrumentalist
    BERT: Singer
    MPNET-Q: Musician
    
    Query: most recently released
    BERT: popular
    MPNET-Q: latest
    
    Query: totalled
    BERT: count
    MPNET-Q: sum
\end{Verbatim}
In addition, the BERT system tends to rank antonyms higher.
\begin{Verbatim}[fontsize=\small]
    Query: min
    BERT: highest
    MPNET-Q: lowest
    
    Query: hilarious
    BERT: erotic
    MPNET-Q: comedy
    
    Query: resigned
    BERT: active
    MPNET-Q: voluntarily terminated
\end{Verbatim}

\subsubsection{Abbreviations \& Short Forms}
All models perform poorly on abbreviations and short forms. BM25 relies on character n-grams to match shortened sub-strings of entities, but fails on acronyms. MPNET-Q is able to identify acronyms of popular entities like universities, countries, etc., perhaps as a result of the fine-tuning phase.
\begin{Verbatim}[fontsize=\small]
    Query: PSU Football 
    BM25: Football
    BERT: UD Arena
    MPNET-Q: Penn State Nittany Lions
             football
    
    Query: Mla
    BM25: Mlabri Language
    BERT: lo
    MPNET-Q: Mlabri Language

    Query: USSR
    BM25: (no result)
    BERT: Czechoslovakia
    MPNET-Q: Soviet Union
\end{Verbatim}

\subsubsection{Numeric Matches}
Among the three systems, BERT performs the worst with numeric entities. It does not handle different numeric representations of the same entity well, leading to random predictions. Fuzzy character matching ensures that BM25 system handles different formats well as long as most of the characters match. MPNET-Q model handles changes in numeric formats the best even when compared against its 
full 
model, with P@1 of 92.8.
\begin{Verbatim}[fontsize=\small]
    Query: 90's
    BM25: 1990s
    BERT: 2010s
    MPNET-Q: 1990s
    
    Query: 5th Avenue
    BM25: 12th Avenue
    BERT: 12th Avenue
    MPNET-Q: 45 Fifth Avenue

    Query: 1775 April 19
    BM25: april 1986
    BERT: 1875-09
    MPNET-Q: 1775-04-19
\end{Verbatim}

\subsubsection{Phonetic Matches}
Often, ASR systems mis-transcribe uncommon words into more common, phonetically similar words. This category tests whether the models are robust to such errors. While the performance of all the models are lacking, BM25 qualitatively provides explainable results due to its reliance on textual similarities when compared to predictions of the 
PTMs.
Typically, EL systems are evaluated on queries that test the models' abilities to match the 4 categories mentioned above. Given the popularity of conversational agents with a speech interface, probing EL models for their phonetic matching capabilities is important.
    
\begin{Verbatim}[fontsize=\small]
    Query: this loud (Liz Laub)
    BM25: Cloud Hu
    BERT: Kevin Upright
    MPNET-Q: Riley Rant
    
    Query: Yale sushi (Xiaoxue Shi)
    BM25: Sakshi Alekar
    BERT: Joshua Frattarola
    MPNET-Q: Sammy Su
\end{Verbatim}

%% file: tables/results_table.tex
\begin{table*}[!ht]
\hspace{-1.5em}
\parbox{1.0\linewidth}{
    \tiny
    \begin{tabular}{l|l||lllll|l||ll||ll}
    \multicolumn{12}{c}{\tiny{\textbf{Results for Entity Linking (Precision@1 / Precision@5)}}} \\ 
    \hline
    & 
    &
    \multicolumn{1}{c}{\multirow{2}{*}{\textsc{mm\_bp}}} & 
    \multicolumn{1}{c}{\multirow{2}{*}{\textsc{complwq}}} & 
    \multicolumn{1}{c}{\multirow{2}{*}{\textsc{mkqa}}} &
    \multicolumn{1}{c}{\multirow{2}{*}{\textsc{acri}}} &
    \multicolumn{1}{c|}{\multirow{2}{*}{\textsc{asr-mis}}} &
    \multicolumn{1}{c||}{\multirow{2}{*}{\textit{Avg.}}} &
    \multicolumn{2}{c||}{\textsc{mm\_bp-word}} &
    \multicolumn{2}{c}{\textsc{mm\_bp-char}} 
    \\
    & 
    &
    & 
    & 
    &
    &
    &
    &
    \multicolumn{1}{c}{\textit{before}} & 
    \multicolumn{1}{c||}{\textit{after}} &
    \multicolumn{1}{c}{\textit{before}} &
    \multicolumn{1}{c}{\textit{after}}
    \\
    \hline
    
    \multicolumn{1}{l|}{\multirow{5}{*}{Baselines}} 
    & \textsc{bm25}
    & 49.5 / 52.6 & 55.1 / 66.0 & 54.2 / 59.6 & 1.1 / 1.5  & \textbf{52.3} / 66.0   & 42.4 / 49.2
    & 41.9 / 54.9  & \underline{33.5} / 41.0 & 46.5 / 60.8  & \textbf{37.6} / 48.9 \\
    
    & \textsc{tfidf}                 
    & 66.7 / 88.3 & 55.7 / 69.9 & 67.3 / \underline{84.0} & 1.0 / 2.1  & 39.4 / \textbf{66.3} & 46.0 / 62.1 
    & 41.1 / 87.5  & \textbf{35.3} / \underline{86.3} & 43.3 / 87.6  & \underline{36.2} / \textbf{87.7} \\ 
    
    & \textsc{Flair}                 
    & 44.1 / 81.7 & 32.8 / 40.2 & 19.7 / 23.8 & 0.1 / 0.3   & 12.9 / 20   & 21.9 / 33.2
    &  17.1 / 81.9  & 11.4 / 79.4 & 20.4 / 82.3  & 12.1 / 75.3 \\ 
    
    & \textsc{Fasttext}                 
    & 60.9 / 89.8 & 37.1 / 47.9 & 24.4 / 30.7 & 6.0 / 11.5    & 4.0 / 7.4     & 26.5 / 37.5
    & 26.0 / 87.4  & 13.5 / 82.4 & 29.4 / 88.7  & 13.5 / 76.1   \\ 
    
    & \textsc{Elmo}                 
    & 57.7 / 84.6 & 46.6 / 58.3 & 26.6 / 32.6 & 1.1 / 1.9   & 6.4 / 10.4  & 27.7 / 37.6
    & 17.5 / 86.3  & 10.4 / 81.5 & 21.6 / 83.3  & 10.9 / 75.1 \\ 
    \hline
    
    \multicolumn{1}{l|}{\multirow{3}{*}{Type-I}} 
    & \textit{bert-base-cased}       
    & 45.9 / 79.4 & 40.2 / 49.2 & 27.2 / 34.3 & 0.6 / 1.2  & 5.7 / 8.9   & 23.9 / 34.6 
    & 15.4 / 82.0  & 8.3 / 75.8  & 18.9 / 78.2  & 9.6 / 73.7 \\
    
    & \textit{roberta-base}          
    & 59.0 / 76.7 & 43.9 / 40.4 & 38.5 / 30.7 & 1.7 / 0.9  & 13.1 / 9.3  & 31.2 / 31.6 
    & 27.4 / 78.8  & 13.5 / 75.3 & 29.4 / 77.1  & 17.2 / 72.3 \\ 
    
    & \textit{mpnet-base}            
    & 31.1 / 74.3 & 25.2 / 30.1 & 21.4 / 23.0 & 0.4 / 0.6  & 12.3 / 14.0 & 18.1 / 28.4 
    & 21.9 / 77.6  & 7.1 / 73.1  & 22.2 / 74.2  & 8.3 / 69.2  \\ 
    \hline
    
    \multicolumn{1}{l|}{\multirow{4}{*}{Type-II}} 
    & \textit{albert-base-v2}       
    & 39.1 / 77.8 & 22.9 / 28.5 & 17.4 / 19.2 & 0.1 / 0.3  & 5.8 / 7.8   & 17.1 / 26.7 
    & 18.2 / 79.9  & 6.6 / 74.4  & 18.1 / 78.3  & 7.0 / 71.7 \\
    
    & \textit{distilbert-base-cased} 
    & 52.8 / 81.1 & 29.5 / 33.3 & 37.7 / 41.0 & 0.6 / 0.7  & 6.5 / 9.5   & 25.4 / 33.1 
    & 15.4 / 83.9  & 9.9 / 74.6  & 19.0 / 80.4  & 11.1 / 72.8 \\
    
    & \textit{distilroberta-base}    
    & 64.6 / 77.7 & 39.6 / 31.9 & 36.1 / 26.5 & 1.5 / 0.7  & 10.6 / 6.5  & 30.5 / 28.7 
    & 25.0 / 80.8  & 14.0 / 74.2 & 26.2 / 76.9  & 16.3 / 73.3 \\ 
    
    & \textit{MiniLM-L6-uncased}     
    & 57.5 / 82.7 & 33.7 / 38.0 & 36.8 / 38.9 & 0.5 / 0.8  & 16.9 / 18.3	& 29.1 / 35.8
    & 29.5 / 85.6  & 9.8 / 75.6  & 30.6 / 81.4  & 11.4 / 73.8 \\
    \hline
    
    \multicolumn{1}{l|}{\multirow{3}{*}{Type-III}} 
    & \textit{all-distilroberta-v1}  & 72.3 / \underline{91.8} & \underline{62.5} / \underline{72.1} & 59.9 / 76.4 & \textbf{11.2 / 23.3} & 42.4 / \underline{62.6} & 49.7 / 65.2 
    & 42.2 / \underline{91.7}  & 24.2 / \textbf{87.3} & 44.3 / 91.4  & 32.2 / \underline{87.5} \\
    
    & \textit{all-mpnet-base-v2}     
    & \underline{75.8} / 91.4 & 62.3 / \underline{72.1} & 57.6 / 73.3 & \underline{6.0} / \underline{11.4} & 43.6 / 57.5 & 49.1 / 61.1 
    & \underline{44.8} / \textbf{92.4}  & 21.6 / 85.4 & 46.3 / 90.9  & 27.5 / 86.3 \\
    
    & \textit{all-MiniLM-L6-v2}      
    & 74.6 / 91.7 & 62.0 / 71.8 & \underline{68.2} / 80.1 & 4.7 / 8.2  & \underline{44.9} / 59.2 & \underline{50.9} / \underline{62.2}
    & \textbf{45.5} / 91.4  & 21.5 / 82.9 & 47.5 / \underline{92.0}  & 27.4 / 84.8 \\ 
    \hline
    
    \multicolumn{1}{l|}{\multirow{2}{*}{Type-IV}} 
    & \textit{all-mpnet-base-v2} (Q) 
    & \textbf{79.4} / \textbf{93.3} & \textbf{63.2} / \textbf{72.4} & \textbf{75.5 / 84.1} & 5.5 / 10.6 & 43.5 / 59.5 & \textbf{53.4} / \textbf{64.0} 
    & \textbf{45.5 / 92.4}  & 22.1 / 85.4 & \underline{46.7} /\textbf{92.5}  & 28.4 / 85.8 \\
    
    & \textit{all-MiniLM-L6-v2} (Q)  
    & 73.0 / 91.0 & 61.2 / 70.9 & 67.3 / 79.6 & 4.1 / 7.4  & 42.3 / 58.1 & 49.6 / 61.4 
    & \underline{44.8} / 91.3  & 20.9 / 83.0 & \textbf{46.8} / 90.9  & 26.5 / 82.3
    \end{tabular}
}
\caption{Evaluation of different pretrained 
transformer models across different datasets (\cref{sec:datasets}). 
The \textit{Avg.} column reports mean precision across different datasets. 
Marked in \textbf{bold} are the best scores \& in \underline{underline} 
are second best.}
\label{tab:results_table}
\end{table*}

%% file: images/bar_plot.tex
\begin{figure}[htb]
\centering
\hbox{
    \hspace{-1.2em}
    \includegraphics[width=0.5\textwidth]{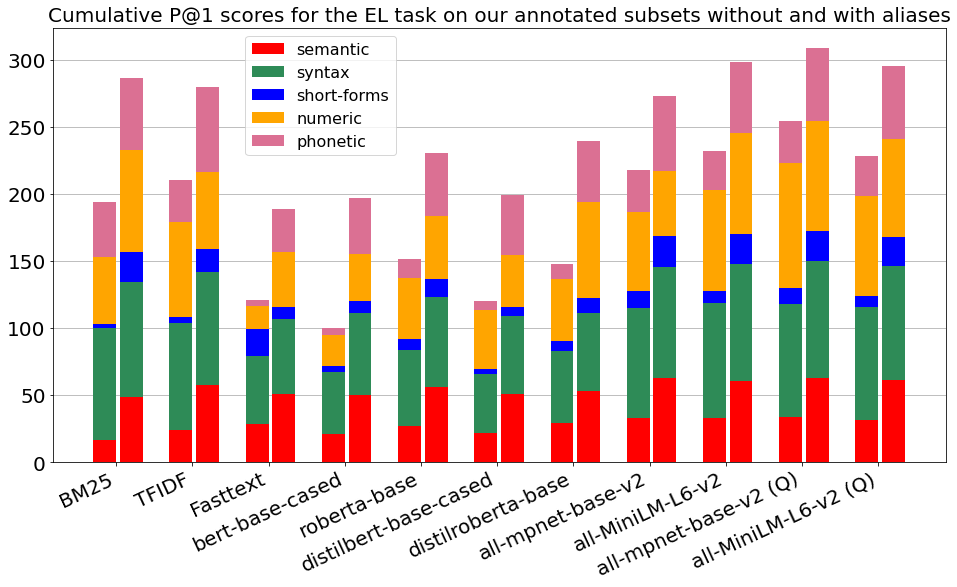}
}
\caption{\label{fig:bar_plot} 
EL results on annotated subset 
of 1.3K test queries, 
annotated across 5 matching criterion. 
For each model,
the first bar 
corresponds to the 
scenario with KBs containing 
only canonical names
whereas for the second,
KBs contain 
aliases in addition 
for disambiguating test queries.
}
\end{figure}

%% file: sections/conclusion.tex
\section{Conclusion}
\label{sec:conclusion}
Given the success of 
PTMs
for various NLP applications~\cite{Rogers_2020}, we evaluate the ability of these models to understand short spans of text for unsupervised entity linking in task-oriented dialog systems by curating a large dataset and comparing their results against traditional n-gram systems. We further analyze the performance of these models across 5 different characteristics- syntactic, semantic, abbreviations \& short-forms, numeric and phonetic matches. Our results demonstrate that these models, when fine-tuned on a semantic similarity task, comprehend syntactic and semantic differences in short phrases better than their other variants. However, their performance is lacking - particularly for abbreviations and queries with speech recognition errors - with the best performing models averaging at 53.4\% P@1 and 64.0\% P@5  across the different datasets. For future work, with the goal of creating a generic model for the unsupervised EL task, we plan to improve these models through task-adaptive fine-tuning techniques with our curated datasets.

%% file: sections/appendix-datasets.tex
\subsection{Dataset Curation}
\label{sec:appendix_datasets}

To conduct a comprehensive evaluation, 
we curate datasets 
for entity linking spanning the following matching criterion--
semantic, syntactic, short-forms, numerals and phonetics.
We utilize both in-house as well as publicly available datasets in our 
evaluation. 
\begin{itemize}
    \item \textsc{mm\_bp}: MindMeld Blueprints\footnote{\href{https://github.com/CiscoDevNet/mindmeld-blueprints}{https://github.com/CiscoDevNet/mindmeld-blueprints}} spanning the domains of food ordering, grocery shopping, video streaming, human resources assistant and banking assistant chatbots.These datasets reflect real-world use-cases with custom entities which are manually annotated by human experts. 
    \item \textsc{mm\_bp-word} and \textsc{mm\_bp-char}: We also curate misspelled versions of the above data to understand the robustness of these models against spelling errors which are common user errors when interacting with a chatbot. We use Neuspell's~\cite{jayanthi-etal-2020-neuspell} word- and character-noising models with default parameter settings to inject misspelling in our test queries. As only a subset of all test queries could be noised, for the misspelling datasets, we benchmark the different models with only the noise-able subset of test queries before and after noising.
    \item \textsc{complwq} and \textsc{mkqa}: To broaden the scope of evaluation to a larger scale, we repurpose Question-Answering (QA) datasets such as ComplexWebQuestions~\cite{talmor18compwebq} and MKQA~\cite{mkqa} for this task. For each answer, we select a subset of answer span aliases with some filtering heuristics and use them as our test queries.
    \item \textsc{acri}: To construct abbreviations and short-forms type matching, we use the Acronym Identification (ACRI) dataset~\cite{veyseh-et-al-2020-what}. The entities obtained from these datasets closely resemble those from free-form and spoken language texts.
    \item \textsc{asr-mis}: Most dialog systems utilize 3rd party speech recognition systems (ASR) that often mistranscribe uncommon words or entities. In such cases, these EL systems need to correct for errors that are beyond textual variations, by utilizing phonetics. To test the phonetic matching capabilities of the models, we include an in-house dataset of ASR mis-transcriptions for person names. To construct this dataset, we randomly sampled ~3K names from the directory of an organization and use Google Speech-to-Text\footnote{\href{https://cloud.google.com/speech-to-text/}{https://cloud.google.com/speech-to-text/}} to collect 10-best ASR transcripts for each sample. 
\end{itemize}

To create the manually annotated sets -- 
\textsc{semantic}, 
\textsc{syntactic},
\textsc{short-forms}, 
\textsc{numerals}
and 
\textsc{phonetics} 
-- we first split the pool of all curated datasets
into 70-30\% ratio at random. We then retain the 70\%
split for training purposes and use the rest 30\%
to sub-sample queries for manual annotation. Retaining
a training split helps to study the impact of using
aliases in entity linking process on each model (Figure 2).

Due to our semi-automated curation strategies, a small portion of test queries might end up with incorrect labels. We plan to address this artifact in future work.

%% file: sections/appendix-experiments.tex
\subsection{Experimental Setup}
\label{sec:appendix_experiments}

\paragraph{MindMeld's Entity Resolution:}
To conduct all our experiments related to Entity Linking,
we use the publicly available open-source library
MindMeld\footnote{\href{https://github.com/cisco/mindmeld}{https://github.com/cisco/mindmeld}}.
MindMeld is a Conversational AI platform for building 
production-quality conversational applications. 
It is a Python-based machine learning framework
which encompasses all of the algorithms and utilities 
required for this purpose.

Due to our deliberate choice to evaluate Entity Linking
step decoupled from Entity Recognition step (to better 
understand the efficacies of PTMs across different matching
criterion),
MindMeld's \textit{Entity Resolution} (ER) 
module\footnote{
\href{https://www.mindmeld.com/docs/userguide/entity\_resolver.html}{
https://www.mindmeld.com/docs/entity\_resolver.html}}
is a perfect choice for our experimental setup. 

The MindMeld ER module provides an easy interface to index Knowledge Bases (KB) with an option of using frameworks like ElasticSearch\footnote{\href{https://www.elastic.co/elasticsearch/}{https://www.elastic.co/elasticsearch/}} for quick retrieval. Various models to be used to extract representations (n-grams or dense representations) can be configured in the config file along with the similarity metric to be used (like BM25, cosine similarity). 

A close alternate choice is 
DeepPavlov's \cite{burtsev-etal-2018-deeppavlov}
Entity Resolution
module\footnote{\href{http://docs.deeppavlov.ai/en/master/features/models/entity\_linking.html}{http://docs.deeppavlov.ai/models/entity\_linking.html}}.

\paragraph{Aliases in Knowledge Bases:}
In experiments wherein KBs contain aliases,
all the aliases' vector representations are treated 
independently and matching to any alias implies matching
to its corresponding canonical title. Meaning, the aliases'
representations are only used to obtain the matching scores but
once obtained, a KB title is assigned the maximum score amongst 
the title's score as well as its aliases'.